\documentclass[conference]{IEEEtran}
\IEEEoverridecommandlockouts
\usepackage{cite}
\usepackage{amsmath,amssymb,amsfonts}
\usepackage{algorithmic}
\usepackage{graphicx}
\usepackage{textcomp}
\usepackage{xcolor}
\usepackage{multirow}
\usepackage{tabularx}

\def\BibTeX{{\rm B\kern-.05em{\sc i\kern-.025em b}\kern-.08em
    T\kern-.1667em\lower.7ex\hbox{E}\kern-.125emX}}
\begin{document}

\title{Memory-guided Network with Uncertainty-based Feature Augmentation for Few-shot Semantic Segmentation\\

\thanks{This work is supported by the Fundamental Research Funds for the Central Universities and the China Scholarship Council.}
}

\author{\IEEEauthorblockN{1\textsuperscript{st} Xinyue Chen}
\IEEEauthorblockA{\textit{Department of Informatics} \\
\textit{King's College London}\\
London, UK \\
xinyue.1.chen@kcl.ac.uk}
\and
\IEEEauthorblockN{2\textsuperscript{nd} Miaojing Shi* \thanks{*Corresponding author}}
\IEEEauthorblockA{\textit{College of Electronic and Information Engineering} \\
\textit{Tongji University}\\
Shanghai, China \\
mshi@tongji.edu.cn}
}

\maketitle

\begin{abstract}
The performance of supervised semantic segmentation methods highly relies on the availability of large-scale training data. To alleviate this dependence, few-shot semantic segmentation (FSS) is introduced to leverage the model trained on base classes with sufficient data into the segmentation of novel classes with few data. FSS methods face the challenge of model generalization on novel classes due to the distribution shift between base and novel classes. To overcome this issue, we propose a class-shared memory (CSM) module consisting of a set of learnable memory vectors. These memory vectors learn elemental object patterns from base classes during training whilst re-encoding query features during both training and inference, thereby improving the distribution alignment between base and novel classes. Furthermore, to cope with the performance degradation resulting from the intra-class variance across images, we introduce an uncertainty-based feature augmentation (UFA) module to produce diverse query features during training for improving the model's robustness. We integrate CSM and UFA into representative FSS works, with experimental results on the widely-used PASCAL-5$^i$ and COCO-20$^i$ datasets demonstrating the superior performance of ours over state of the art.
\end{abstract}


\section{Introduction}
Semantic segmentation is a fundamental computer vision task. Since the introduction of fully convolutional network (FCN) \cite{long2015fully}, significant advances in semantic segmentation \cite{2015unet,2017pspnet,2022mask2former,yu2015multi,chen2017deeplab,fu2019dual,huang2019ccnet,tao2020hierarchical} have been achieved. Despite the great progress, the supervised learning pipeline suffers from performance degradation when facing novel classes with few examples. To tackle this issue, few-shot semantic segmentation (FSS) has been proposed to develop models that can effectively segment objects or regions based on a very limited amount of annotated training data.

FSS aims to learn a model on base classes that can quickly adapt to novel classes with a few support samples, while a distribution shift between base and novel classes naturally arises in this process. In order for the model to achieve the generalizability on novel classes, existing FSS approaches \cite{dong2018few, wang2019panet,liu2020prototype,zhang2021SCL,zhang2022mfnet,liu2022ntrenet,yang2023mianet,zhang2021cyc,zhang2022mm-former} adopt the episodic learning that organizes the training data into episodes, each consisting of a query image and few support images of the same base class to simulate the few-shot scenario. They normally utilize the information contained in annotated support images of a certain class to perform pixel-wise classification of this class on the query image via non-parametric similarity measurement \cite{wang2019panet,liu2020prototype,dong2018few} or parametric decoding \cite{zhang2021SCL,zhang2022mfnet,liu2022ntrenet,yang2023mianet,zhang2021cyc,zhang2022mm-former}. The backbone architecture can be convolutional neural network (CNN)-based \cite{dong2018few,wang2019panet,liu2020prototype,zhang2021SCL,zhang2022mfnet,liu2022ntrenet,yang2023mianet} or transformer-based  \cite{zhang2021cyc,zhang2022mm-former}.
In light of the disturbance of base classes for the prediction of novel classes, BAM \cite{lang2022bam} recently introduces an additional base learner which explicitly segments objects of base classes in the query image so that regions of base classes are effectively eliminated in the meta learner's prediction. This is later adopted in \cite{peng2023hierarchical} as well.

Despite the performance improvement achieved in existing FSS approaches, the underlying distribution shift between base and novel classes still hinders the knowledge transfer from base to novel classes. To alleviate this issue, in this work we introduce the class-shared memory (CSM) module to align the feature representations between base and novel classes. This module consists of a set of memory vectors that learn and store the elemental information (\textit{e.g.}, local patterns of objects) shared across base classes during training; they are used to re-encode features of query images during both training and inference. Consequently, the representations of novel classes become aligned with those of the base classes, enhancing the model's ability to recognize novel classes.

Besides the distribution shift challenge, the intra-class variance (\textit{e.g.}, appearance, pose, scale, viewpoint, environment) between the support and query images also poses a challenge in FSS. To mitigate the negative impact of intra-class variance, previous methods \cite{2020pfenet,li2021asgnet,lang2022bam} utilize various data augmentation techniques, including flip, rotation, scale, crop, \textit{etc.} to enrich the input images. These data augmentation techniques operate in the image space. \cite{hu2022adversarial,chen2019multi,wang2023feature,li2021simple} instead leverage adversarial training, language information, inter-class relationship, \textit{etc.} to augment data in the feature space. 
%but they cannot estimate the probability of augmented features using explicit distributions during the augmentation.  
%and the relationship between support and query features during training.
%Unlike these approaches, our method employs uncertainty estimation to obtain new feature statistics, thereby augmenting features.} 
%To overcome these shortcomings, 
In this work, 
we design a new uncertainty-based feature augmentation (UFA) module. 
%to augment the training data in the feature space by considering the correlation between both support and query features. 
%This module first measures the uncertainty of feature statistics for query and support images; then mixes statistics randomly drawn from the distributions of uncertain feature statistics of query and support images; and finally produces synthesized feature representations for query images using mixed feature statistics during training.
It first measures the uncertainty of feature statistics for query and support features; then produces diverse feature representations for query images during training by combining the feature statistics of both query and support features. 
By using this module, our model becomes more robust to the intra-class variance. 

Overall, the contribution of this paper is threefold:
\begin{itemize}
    \item To address the class distribution shift in FSS, we propose the CSM module that re-encodes the query features of base and novel classes so as to align their representations.
    \item To cope with the intra-class variance in FSS, we propose the UFA module that augments query features for each class by mixing feature statistics drawn from the uncertainty distributions of both query and support features. %based on the feature uncertainty of both query and support images.
    \item We extensively evaluate our method, memory-guided network with uncertainty-based feature augmentation (MENUA), on two public datasets, PASCAL-5$^i$ and COCO-20$^i$. The experimental results demonstrate that our method clearly improves state of the art.
\end{itemize}

\section{Method}\label{subsec2}

\subsection{Problem Definition}\label{subsec21}

For few-shot semantic segmentation, the entire dataset consists of a training set $D_{train}$ and a test set $D_{test}$, where $D_{train}$ has $C_{base}$ base classes with plenty of annotated images for each class; $D_{test}$ has $C_{novel}$ novel classes with few annotated images per class. These two class sets are disjoint, \textit{i.e.},  $C_{base} \cap C_{novel} = \emptyset$. 

Most current FSS methods \cite{2020pfenet,li2021asgnet,lang2022bam} follow the paradigm of episodic learning \cite{vinyals2016matching}. For the 1-way $K$-shot setting, an episode is represented by a support set and a query set. The support set consisting of $K$ samples (\textit{i.e.}, $K$-shot) of a certain class $c$ (\textit{i.e.}, 1-way) is denoted as $\mathcal{S}=\{\mathcal{S}_1, \mathcal{S}_2, ..., \mathcal{S}_K\}$. The $i$-th support sample $\mathcal{S}_i$ is formed by a support image $x_i^\text{s}$ and the corresponding label $ y_i^\text{s}$. The query set including one sample of the same class $c$ is represented by $\mathcal{Q} = \{x^\text{q}, y^\text{q}\}$.
%Both query and support images in each episode belong to a specific category. 
The query-support pair $\{x^\text{q},\mathcal{S}\}$ is provided as known information, while the ground truth $y^\text{q}$ of the query image is only available during training.
\begin{figure*}[!t]
	\centering
	\includegraphics[width=1.0\textwidth]{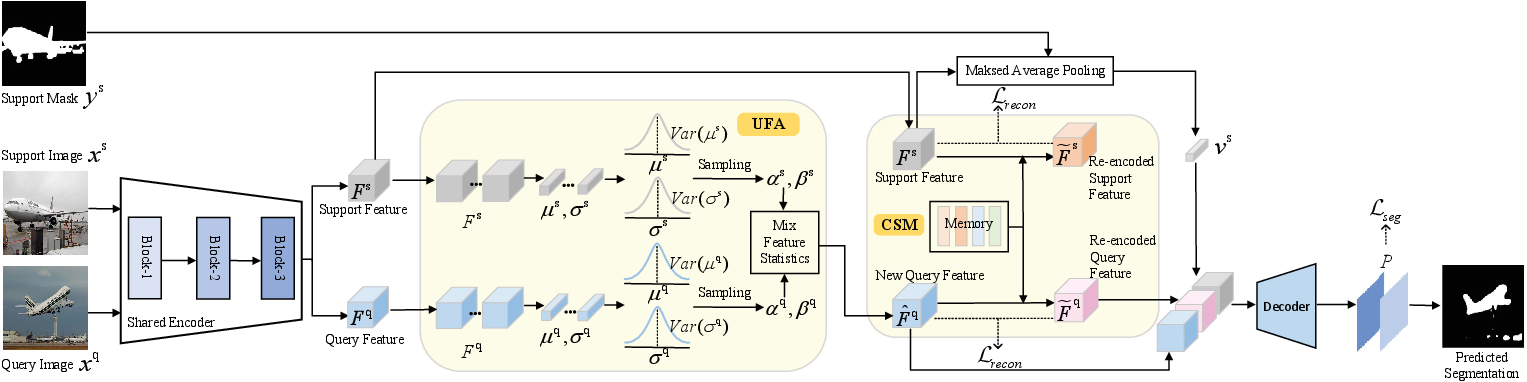}
	\caption{Overview of our MENUA. The uncertainty-based feature augmentation (UFA) module and class-shared memory (CSM) module are shown in the boxes with yellow backgrounds.}
	\label{fig:overview}
\end{figure*}

\subsection{Overview}\label{subsec22}

\figurename \ref{fig:overview} gives an overview of the proposed method,  MENUA. Our network consists of two new modules, \textit{i.e.}, uncertainty-based feature augmentation (UFA, Section \ref{subsec23}) module and class-shared memory (CSM, Section \ref{subsec24}) module, which, without loss of generality, can be easily implanted in existing state of the art architectures (see Section \ref{subsec31}). Specifically, given a query-support pair $\{x^\text{q},\mathcal{S}\}$, we utilize a shared encoder to extract support and query features, denoted by $F^\text{q}$ and $F^\text{s}$. After applying the masked average pooling on $F^\text{s}$, the support prototype $v^\text{s}$ is obtained. For multiple shots, we average the prototypes extracted from all support images. 
The UFA module is introduced to diversify the query feature derived from the encoder, obtaining the diversified version $\widehat{F}^\text{q}$. Next, the CSM module, storing the elemental information shared among classes is used to re-encode the $\widehat{F}^\text{q}$ %query feature
into $\widetilde{F}^\text{q}$. After that, $\widetilde{F}^\text{q}$ and $\widehat{F}^\text{q}$, together with $v^\text{s}$, are concatenated and fed into the decoder (see Section \ref{subsec31}) to yield the prediction result $P$. The details of the training and inference procedures can be found in Section \ref{subsec25}.

\subsection{Uncertainty-Based Feature Augmentation}\label{subsec23}

Since there exists intra-class variance within each class, we aim to increase the sample diversity of each class in the feature space during training
%generate diverse variants of features during training 
so that the model is more robust to the input variance. We propose the UFA module (\figurename \ref{fig:overview}) to diversify query features extracted from the encoder by adjusting their feature statistics, \textit{i.e.}, the feature mean and standard deviation. To implement this module, we hypothesize that each feature statistic follows a Gaussian distribution rather than being a deterministic value \cite{kendall2017uncertainties,li2022dsu}. 
After sampling mean and standard deviation values from their respective distributions, we use them to synthesize new feature representations for query images. This process is described below.    

\subsubsection{Uncertainty estimation} 
The essential idea is first to perform uncertainty estimation on feature statistics to obtain their probabilistic distributions; next, samples are drawn from these probabilistic distributions to generate new features with semantic variations. Specifically, we calculate the feature mean and standard deviation across the spatial dimensions of each channel map of the feature. Given the feature $F \in \mathbb{R}^{H \times W \times C}$, its mean and standard deviation are denoted as ${\mu} \in \mathbb{R}^{1 \times C}$ and ${\sigma} \in \mathbb{R}^{1 \times C}$, respectively. Then, we estimate uncertainty for feature statistics by computing the variances of ${\mu}$ and ${\sigma}$ for each mini-batch, because \cite{li2022dsu,wang2019implicit} show that the variances between features implicitly reflect semantic variations between features. The formulation for the support and query images is similar, we only write out the uncertainty estimation for the query feature:
\begin{equation}
\label{eq:variancemean1}
	Var({\mu}^\text{q})=\frac{1}{N_{bs}}\sum\nolimits_{i=1}^{N_{bs}}({\mu}_{i}^\text{q} - \mathbb{E}[{\mu}^\text{q}])^2
\end{equation}
\begin{equation}
\label{eq:variancesigma1}
Var({\sigma}^\text{q})=\frac{1}{N_{bs}}\sum\nolimits_{i=1}^{N_{bs}}({\sigma}_i^\text{q}-\mathbb{E}[{\sigma}^\text{q}])^2
\end{equation}
\noindent where ${\mu}^\text{q}$ and ${\sigma}^\text{q}$ represent feature statistics of the query image. $Var({\mu}^\text{q})$ and $ Var({\sigma}^\text{q})$ denote the variances of the query feature statistics, and $N_{bs}$ is the number of samples in a mini-batch. Similarly, the variances of the support feature statistics are represented by $Var({\mu}^\text{s})$ and $Var({\sigma}^\text{s})$.

\subsubsection{Gaussian re-parameterization} 
After the uncertainty estimation, we establish the probabilistic distribution of each feature statistic by fitting the Gaussian function \cite{kendall2017uncertainties}. The new support and query feature statistics can be randomly sampled from the Gaussian distributions according to the re-parameterization technique \cite{kingma2013auto}. The formulation for the support and query feature statistics is similar, below we write out the re-parameterization for query feature statistics: 
\begin{equation}\label{eq:mu}
	{\alpha}^\text{q} = \mu^\text{q} + \epsilon_{\mu}^\text{q}Var({\mu}^\text{q}),  \quad  \epsilon_{\mu}^\text{q} \sim \mathcal{N}(0, 1)
\end{equation} 
\begin{equation}\label{eq:sigma}
	{\beta}^\text{q} = \sigma^\text{q} + \epsilon_{\sigma}^\text{q}Var({\sigma}^\text{q}),  \quad  \epsilon_{\sigma}^\text{q} \sim \mathcal{N}(0, 1)
\end{equation}  
\noindent where $\alpha^\text{q} \sim \mathcal{N}(\mu^\text{q}, Var({\mu}^\text{q}))$ and $\beta^\text{q} \sim \mathcal{N}(\sigma^\text{q}, Var({\sigma}^\text{q}))$ are the new mean and standard deviation of query feature respectively. Similarly, we can also obtain $\alpha^\text{s}$ and $\beta^\text{s}$. 

\subsubsection{Feature augmentation} We obtain the variant of the query feature by replacing its original feature statistics with mixed ones. The process of generating mixed feature mean $\widehat{\alpha}^\text{q}$ and standard deviation $\widehat{\beta}^\text{q}$ can be formulated as:
\begin{equation}\label{eq:mixalpha}
	\widehat{\alpha}^\text{q} = \lambda {\alpha}^\text{q}+(1-\lambda)\alpha^\text{s};~~\widehat{\beta}^\text{q} = \lambda {\beta}^\text{q}+(1-\lambda)\beta^\text{s}
\end{equation}
\noindent $\lambda$ is the weight drawn from the Beta distribution, $\lambda \sim B(\gamma,\gamma)$ with $\gamma=0.1$. $({\alpha}^\text{q}, {\alpha}^\text{s})$ are new feature means, and $({\beta}^\text{q}, {\beta}^\text{s})$ are new feature standard deviations sampled from their probabilistic distributions.

The process of generating the variant of the query feature can be formulated as follows:
\begin{equation}\label{eq:newfq}
	\widehat{F}^\text{q} = \widehat{\beta}^\text{q}\frac{F^\text{q}-{\mu}^\text{q}}{\sigma^\text{q}} + \widehat{\alpha}^\text{q}
\end{equation}
\noindent where $F^\text{q}$ and $\widehat{F}^\text{q}$ represent the original query feature and the new query feature with mixed feature statistics, respectively. ${\mu}^\text{q}$ and ${\sigma}^\text{q}$ are the feature mean and standard deviation of $F^\text{q}$. $\frac{F^\text{q}-{\mu}^\text{q}}{\sigma^\text{q}}$ is the standardization operation performed on the query feature to remove its original feature statistics. Then, the mixed feature mean $\widehat{\alpha}^\text{q}$ and standard deviation $\widehat{\beta}^\text{q}$ are applied to generating the new query feature. We combine feature statistics of query and support features in (\ref{eq:mixalpha}) since the differences between query and support feature statistics imply potential intra-class variance. 
We apply feature augmentation on query features only but not on support features because support features are used to provide query features with accurate object class information while augmenting support features can result in inaccurate guidance for the query.

\subsection{Class-Shared Memory for Feature Reconstruction}\label{subsec24}

Due to the commonly occurring distribution shift between base and novel classes, the model encounters obstacles when recognizing novel classes. While there exist semantic differences among based and novel classes, they also consist of shared elemental information, such as local patterns and traits of objects. To alleviate the adverse effect of the distribution shift, we use the memory network to extract class-agnostic elemental information from base classes during training and store this information in memory vectors. 
These learned memory vectors are used to re-encode query features during both training and inference. In this way, we successfully narrow the gap between the feature representations of base and novel classes. The process is specified below.

\subsubsection{Feature reconstruction} 
The proposed CSM is composed of a set of learnable memory vectors, denoted as $\mathcal{M} = \{m_k\}_{k=1}^N$. Taking the diversified query feature $\widehat{F}^\text{q} \in \mathbb{R}^{H \times W \times C}$ as an example, for each query vector, \textit{e.g.}, $f_{j}^\text{q} \in \mathbb{R}^{1 \times C}$ at $j$-th pixel in $\widehat{F}^\text{q}$, we compute its similarity to every of memory vectors $\{m_k\}$ and then use a Softmax operation over these similarity values to obtain a set of weights for calculating the weighted sum of all memory vectors, resulting into the new query vector $\widetilde{f}_{j}^\text{q}$. Finally, the re-encoded query features $\widetilde{F}^\text{q}$ are obtained by iteratively processing every pixel vector in $\widehat{F}^\text{q}$.  
\begin{equation}\label{eq:sim}
	\widetilde{f}_{j}^\text{q} = \sum\nolimits^{N}_{k=1} m_k \cdot Softmax(f_{j}^\text{q} m_k^\top) 
\end{equation}
\noindent where $N$ is the number of memory vectors. $\widetilde{f}_{j}^\text{q}$ is the vector at $j$-th pixel in the re-encoded query feature $\widetilde{F}^\text{q}$. 

The re-encoded support feature $\widetilde{F}^\text{s}$ is obtained by the same process as $\widetilde{F}^\text{q}$. Both are used to compute the reconstruction loss (see Section \ref{subsec25}) during training. 

\subsubsection{Memory update} 
Memory vectors in $\mathcal M$ are learnable and are updated by minimizing the feature reconstruction loss, specified below in Section \ref{subsec25}. As shown in \figurename \ref{fig:overview}, we use the reconstruction loss derived from the re-encoded query and support features to train these memory vectors, but the support prototype $v^s$ is provided by the original support feature $F^\text{s}$ rather than the re-encoded support feature $\widetilde{F}^\text{s}$. Additionally, all memory vectors are used to re-encode the query feature during both training and testing. 

Notice memory network has been used in previous FSS works~\cite{ramalho2019adaptive,he2020memory,wu2021learning}. For instance, \cite{ramalho2019adaptive,he2020memory} adopt episodic memory that is destroyed when a new episode arrives. \cite{wu2021learning} adopts the memory that exists across episodes and leverages memory vectors to get the activation maps of query and support features. Different from them, we initialize memory vectors based on representations of base classes and utilize them to re-encode query features for mask generation.

%can you add the some sentences here to say the comparison to other related works use memory vectors in FSS and differences. 
%randomly initialized and then learn to store elemental information during base-learner training. These memory vectors are updated by minimizing the reconstruction loss during meta-learner training (Section \ref{subsec43}). $\mathcal M$ is utilized to reconstruct query features during both meta-learner training and testing.}

\subsection{Training and Inference}\label{subsec25}
In this section, we introduce losses for training as well as details for inference.

\subsubsection{Feature reconstruction loss} 
We introduce a feature reconstruction loss for the optimization of CSM. It lets the re-encoded features ($\widetilde{F}^\text{s}$ and $\widetilde{F}^\text{q}$) become close to their corresponding original features ($F^\text{s}$ and $\widehat{F}^\text{q}$), which prevents them from collapsing. We take the example of $\widehat{F}^\text{q}$ and $\widetilde{F}^\text{q}$, to show the process. First, to restrict these memory vectors to learn the foreground information related to targets, we mask the background of $\widehat{F}^\text{q}$ and $\widetilde{F}^\text{q}$ using the ground truth mask of the query image. Then, we compute the correlation matrix $\mathcal{C}\in \mathbb{R}^{HW\times HW}$ by multiplying the masked $\widehat{F}^\text{q}$ and masked $\widetilde{F}^\text{q}$. We define the reconstruction loss $\mathcal{L}_{recon}$ as a masked cross-entropy loss, with the goal of maximizing the diagonal elements of $\mathcal{C}$ (\textit{i.e.}, $\mathcal{C}_{diag}$), hence making the foreground of $\widehat{F}^\text{q}$ and $\widetilde{F}^\text{q}$ highly correlated:
\begin{equation} \label{loss}
	\mathcal{L}_{recon} = \sum\nolimits_{diag=1}^{HW}({y'} \times CE(\mathcal{C}_{diag}))
\end{equation}
\noindent where $CE$ is the cross-entropy loss. $y{'} \in \mathbb{R}^{1\times HW}$ is the flattened ground truth mask of the support/query image, indicating which elements in $\mathcal{C}_{diag}$ should be involved to calculate this loss. We use the masked cross-entropy loss instead of the standard cross-entropy loss \cite{wu2021learning} since there exist zero elements (\textit{i.e.}, background pixels) in $\mathcal{C}_{diag}$ and they remain unchanged during training. The masked cross-entropy loss is designed to only sum the loss values from foreground regions.

\subsubsection{Training} 
The total training loss consists of the feature reconstruction loss and the typical segmentation loss: 
$\mathcal{L}_{total} = \mathcal{L}_{seg} + \mathcal{L}_{recon}$. 
The segmentation loss $\mathcal{L}_{seg}$ is a binary cross-entropy loss to supervise the predicted segmentation results.  

\subsubsection{Inference} In the inference stage, we utilize the memory vectors in CSM to re-encode query features while UFA is used only in the training stage.

\section{Experiments}\label{sec5}
\begin{table}[t]
\centering
\small
\caption{Performance comparison on PASCAL-5$^i$.}\label{tab:overall_pascal}
\begin{tabular}{cccc}
\hline
\multicolumn{1}{c}{Backbone}  & \multicolumn{1}{|c|}{Method} & \multicolumn{1}{c|}{1-shot}   & \multicolumn{1}{c}{5-shot} \\ \hline
%\multicolumn{3}{c}{mIoU}                                                                                                                                                         \\ \hline
\multicolumn{1}{c}{\multirow{8}{*}{ResNet50}}  & \multicolumn{1}{|c|}{CyCTR \cite{zhang2021cyc}}                  & \multicolumn{1}{c|}{64.20}           &  65.60  \\
                                               & \multicolumn{1}{|c|}{MM-Former \cite{zhang2022mm-former}}                & \multicolumn{1}{c|}{63.30}     &  64.90  \\ 
                                               & \multicolumn{1}{|c|}{NTRENet \cite{liu2022ntrenet}}                  & \multicolumn{1}{c|}{64.20}           &  65.70  \\
                                               & \multicolumn{1}{|c|}{MIANet \cite{yang2023mianet}}                  & \multicolumn{1}{c|}{68.72}           &  71.59  \\ \cline{2-4}
                                               & \multicolumn{1}{|c|}{BAM \cite{lang2022bam}}                    & \multicolumn{1}{c|}{67.81}      & 70.91   \\
                                               & \multicolumn{1}{|c|}{BAM + MENUA}                    & \multicolumn{1}{c|}{${\textbf{69.24}}$}  & ${\textbf{72.07}}$ \\
                                               & \multicolumn{1}{|c|}{HDMNet \cite{peng2023hierarchical}}                       & \multicolumn{1}{c|}{69.40}      & 71.80   \\
                                               & \multicolumn{1}{|c|}{HDMNet + MENUA}                & \multicolumn{1}{c|}{${\textbf{70.22}}$}  & $\textbf{72.89}$ \\ \hline
\end{tabular}
\end{table}
% \usepackage{multirow}
\begin{table}[t]
\centering
\small
%\resizebox{\textwidth}{!}{
\caption{Performance comparison on COCO-20$^i$.}\label{tab:overall_coco}
\begin{tabular}{cccc}
\hline
\multicolumn{1}{c|}{Backbone}  & \multicolumn{1}{c|}{Method} & \multicolumn{1}{c|}{1-shot}   & \multicolumn{1}{c}{5-shot} \\ \hline
%\multicolumn{3}{c}{mIoU}                                                                                                                                                         \\ \hline
\multicolumn{1}{c}{\multirow{8}{*}{ResNet50}} & \multicolumn{1}{|c|}{CyCTR \cite{zhang2021cyc}}                  & \multicolumn{1}{c|}{40.30}           &  45.60  \\
                                              & \multicolumn{1}{|c|}{MM-Former \cite{zhang2022mm-former}}                & \multicolumn{1}{c|}{44.20}      &  48.40  \\
                                              & \multicolumn{1}{|c|}{NTRENet \cite{liu2022ntrenet}}                  & \multicolumn{1}{c|}{39.30}           &  40.30  \\
                                              & \multicolumn{1}{|c|}{MIANet \cite{yang2023mianet}}                  & \multicolumn{1}{c|}{47.66}           &  51.65  \\ \cline{2-4}
                                              & \multicolumn{1}{|c|}{BAM \cite{lang2022bam}}                     & \multicolumn{1}{c|}{46.23} & 51.16  \\
                                              & \multicolumn{1}{|c|}{BAM + MENUA}                   & \multicolumn{1}{c|}{${\textbf{47.74}}$}  & ${\textbf{52.25}}$  \\ 
                                              & \multicolumn{1}{|c|}{HDMNet \cite{peng2023hierarchical}}                       & \multicolumn{1}{c|}{50.00}      & 56.00   \\
                                              & \multicolumn{1}{|c|}{HDMNet + MENUA}                & \multicolumn{1}{c|}{${\textbf{50.94}}$}  & ${\textbf{57.08}}$ \\ \hline
\end{tabular}
%\vspace{-2mm}
\end{table}
\begin{figure}[t]%
	\centering
	\includegraphics[width=0.95\linewidth]{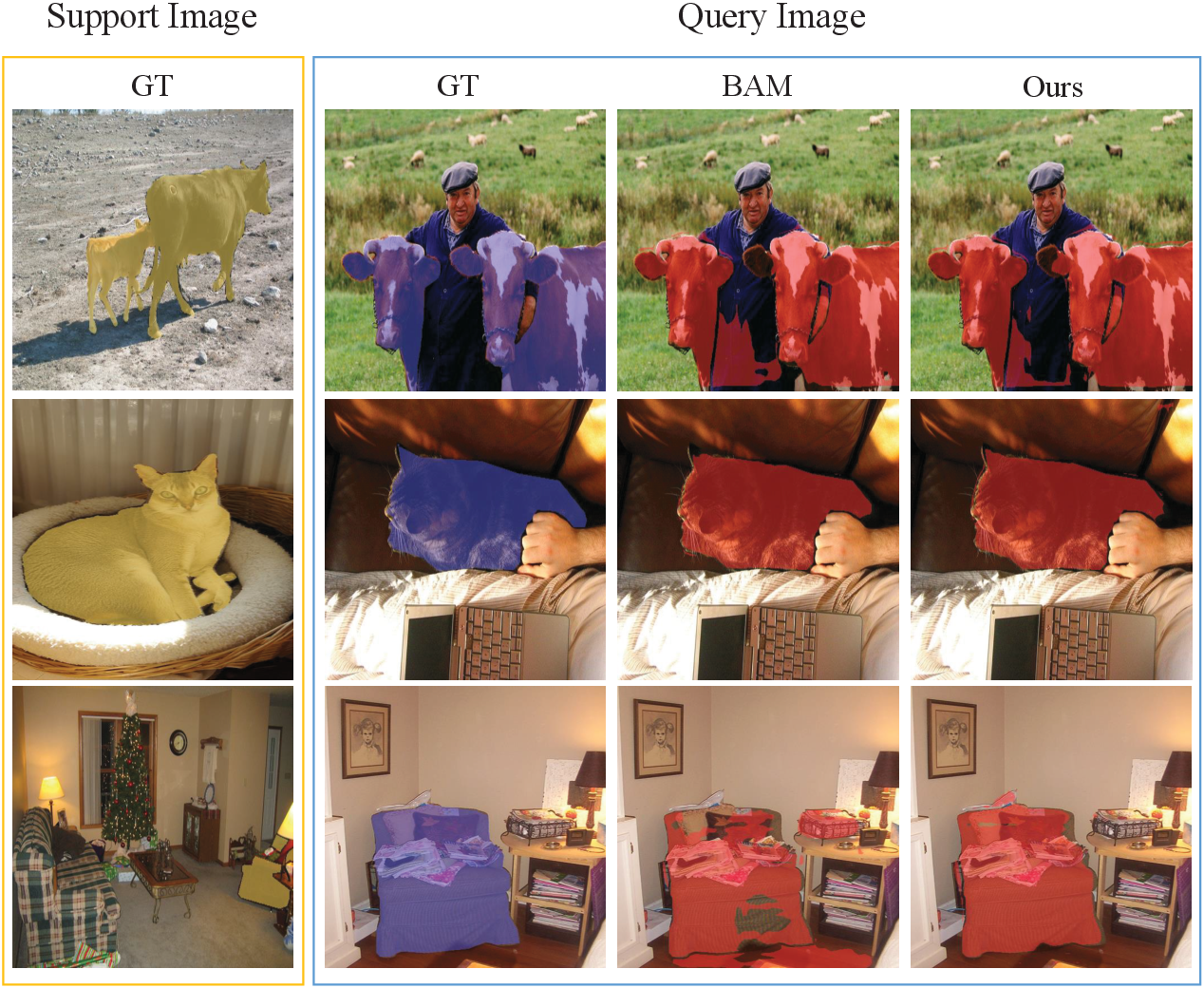}
	\caption{The examples of the segmentation results for BAM \cite{lang2022bam} and our proposed MENUA on PASCAL-5$^i$.}
	\label{fig:prediction}
\end{figure}

\begin{table}[t]
\begin{minipage}[c]{0.54\linewidth}
\centering
\small
\caption{Ablation study on our proposed modules.}
\begin{tabularx}{\linewidth}{>{\centering\arraybackslash}p{12mm}>{\centering\arraybackslash}p{6mm}>{\centering\arraybackslash}p{6mm}|>{\centering\arraybackslash}p{7mm}}
\hline
Backbone &  CSM   & UFA  & mIoU      \\ 
\hline
\multirow{4}{*}{ResNet50} &             &            & 67.42         \\
			& $\checkmark$  &              & 68.58                    \\
			&               & $\checkmark$  & 68.27                    \\
			& $\checkmark$  & $\checkmark$  & \textbf{69.24}    \\
			\hline
\end{tabularx}
%\vspace{-2mm}
\label{tab:abl_overall}
\end{minipage}
\hspace{2mm}
\begin{minipage}[c]{0.4\linewidth}
\centering
\small
\caption{Ablation study on feature augmentation in UFA.}
\begin{tabular}{c|c}
\hline
Method & mIoU           \\ \hline
%Flip \cite{lang2022bam} & 68.70 \\
CutMix \cite{yun2019cutmix} & 67.45 \\
UFA (Q+S)    & 67.91          \\
UFA (Q)  & \textbf{69.24} \\ \hline
\end{tabular}
%\vspace{-2mm}
\label{tab:abl_ufa}
\end{minipage}
\end{table}

\begin{table}[t]
\begin{minipage}[c]{0.5\linewidth}
\centering
%\vspace{-2mm}
\small
\caption{Ablation study on the way of feature reconstruction.}
\begin{tabular}{c|c}
\hline
Method  & mIoU  \\ \hline
CSM ($k$=1) & 67.67 \\
CSM ($k$=5) & 68.46      \\
CSM ($k$=all)  & \textbf{69.24} \\ \hline
\end{tabular}
%\vspace{-2mm}
\label{tab:abl_fr}
\end{minipage}
\hspace{5mm}
\begin{minipage}[c]{0.42\linewidth}
\centering
%\vspace{-2mm}
\small
\caption{Ablation study on the feature reconstruction loss.}
\begin{tabular}{c|c}
\hline
Loss & mIoU           \\ \hline
w/o $\mathcal{L}_{recon}$    & 67.81      \\
w $\mathcal{L}_{recon}$  & \textbf{69.24} \\ \hline
\end{tabular}
%\vspace{-2mm}
\label{tab:abl_loss}
\end{minipage}
\end{table}

\begin{table}[hbt!]
\begin{minipage}[c]{0.45\linewidth}
\centering
%\vspace{-2mm}
\small
\caption{Ablation study on the number of memory vectors in CSM.}
\begin{tabular}{c|c}
\hline
Number & mIoU \\ \hline
20     & 67.99 \\
30     & 68.21 \\
50     & \textbf{69.24}  \\
100 & 68.93  \\
200     &  68.80  \\ \hline
\end{tabular}
%\vspace{-2mm}
\label{tab:abl_memory}
\end{minipage}
\hspace{9mm}
\begin{minipage}[c]{0.42\linewidth}
\centering
%\vspace{-2mm}
\small
\caption{Ablation study on inserted positions of UFA.}
\begin{tabular}{c|c}
\hline
Position & mIoU \\ \hline
B0+1     & 69.08  \\
B1+2     & \textbf{69.24}  \\
B2+3     &  68.98      \\
B0+1+2   & 68.84  \\
B1+2+3   & 69.04  \\ 
B0+1+2+3 & 69.10 \\ \hline
\end{tabular}
%\vspace{-2mm}
\label{tab:position}
\end{minipage}
\end{table}
\subsection{Setup}\label{subsec31}
\subsubsection{Datasets} 
We evaluate our MENUA on two public datasets, PASCAL-5$^i$ \cite{shaban2017one} and COCO-20$^i$ \cite{nguyen2019feature}. PASCAL-5$^i$ includes 20 categories and COCO-20$^i$ includes 80 categories, and their object categories are evenly split into four folds. We conduct experiments in a cross-validation manner: when one fold is regarded as the test set, the other three folds serve as the training set. This process is repeated for four rounds for traversing all folds for testing, and we report the average result over these rounds.

\subsubsection{Implementation details} 
While our proposed MENUA is easily plugged to previous FSS models, we mainly build it upon two state of the arts, \textit{i.e.}, BAM \cite{lang2022bam} and HDMNet \cite{peng2023hierarchical}. Both BAM and HDMNet use ResNet50 as the backbone encoder; while BAM uses ASPP \cite{chen2017deeplab} as the decoder and HDMNet uses PPM \cite{2017pspnet}. All training settings (\textit{e.g.}, optimizer, learning rate, batch size, epochs, and data augmentation strategy) are the same to BAM/HDMNet. In our experiments, UFA modules are inserted after the Block-$1$ and Block-$2$ of ResNet50 to diversify query features. Additionally, memory vectors in CSM are randomly initialized, there are $50$ vectors, each with dimensions of 256.  
%Notably, BAM and HDMNet adopt the two-stage training, \textit{i.e.}, base-learner training and meta-learner training. For the first stage, we randomly initialize the memory vectors in CSM. For the second stage, we use the memory vectors obtained during base-learner training to initialize the memory vectors in meta-learner training.}
Experiments are implemented in the PyTorch with NVIDIA A100 GPU.

\subsubsection{Evaluation metric} 
Following the previous work \cite{2020pfenet}, we adopt the mean intersection-over-union (mIoU) as the evaluation metric, $mIoU = \frac{1}{C_{novel}}\sum\nolimits^{C_{novel}}_{i=1} IoU_i$, where $C_{novel}$ is the number of novel classes in each fold (\textit{e.g.}, $C_{novel}=5$ for PASCAL-$5^i$ and $C_{novel}=20$ for COCO-$20^i$) and $IoU_i$ represents the intersection-over-union between the predicted segmentation results and ground truth masks for the $i$-th novel class.

\subsection{Comparison with State of the Art}\label{subsec32} 

\subsubsection{Quantitative comparisons} 
Table \ref{tab:overall_pascal} and Table \ref{tab:overall_coco} illustrate the comparisons between FSS models and our MENUA on PASCAL-5$^i$ and COCO-20$^i$. Our MENUA improves the state of the art performance in both 1-way 1-shot and 5-shot settings. Specifically, on PASCAL-5$^i$, our method improves BAM \cite{lang2022bam} by 1.43$\%$ and 1.16$\%$ mIoU and improves HDMNet \cite{peng2023hierarchical} by 0.82$\%$ and 1.09$\%$ mIoU. The performance of our method outperforms previous CNN-based methods (\textit{i.e.}, NTRENet \cite{liu2022ntrenet} and MIANet \cite{yang2023mianet}) and transformer-based methods (\textit{i.e.}, CyCTR \cite{zhang2021cyc} and MM-Former \cite{zhang2022mm-former}). Similarly, experimental results on COCO-20$^i$ show that our method brings improvements over our baselines (\textit{i.e.}, BAM and HDMNet) and achieves the state of the art performance. For the sake of space; we refer readers to the supplementary material for the detailed result in every test fold of PASCAL-5$^i$ and COCO-20$^i$. 

\subsubsection{Qualitative comparison}
We show the qualitative comparison between ours and BAM in \figurename \ref{fig:prediction}. Specifically, the ground truth masks for support and query images are indicated in yellow and blue, respectively; while the predicted segmentation results for BAM and our method are in red. Our method (4-th column) performs better than BAM (3-rd column) in recognizing details (\textit{e.g.}, the ears of the cow and cat) and suppressing distracting objects.

\subsection{Ablation Study}\label{subsec33}

We select BAM \cite{lang2022bam} as the baseline and report the average mIoU across 4 folds on PASCAL-5$^i$ in the 1-way 1-shot.
%in Table \ref{tab:abl_overall}, \ref{tab:abl_init}, \ref{tab:number}, \ref{tab:position} are .}

\subsubsection{CSM and UFA modules} 
We evaluate the performance of CSM and UFA separately. As shown in Table \ref{tab:abl_overall}, our model, equipped with CSM and UFA modules, achieves the best performance with 69.24$\%$ in mIoU. Notably, a decrease of 0.97$\%$ in mIoU is observed when removing the CSM module, while a reduction of 0.66$\%$ in mIoU by removing the UFA module. This validates the effectiveness of our proposed CSM and UFA for FSS.

\subsubsection{Feature augmentation in UFA} 
%(original: We conduct experiment to investigate the application of our UFA to only query images (UFA (Q), default) or both query and support images (UFA (Q+S)). Table \ref{tab:abl_ufa} presents the results: augmenting support features results in performance degradation. This is consistent with our analysis in Section \ref{subsec23}.) 
We conduct experiments to evaluate the effectiveness of our UFA. First, we investigate the application of our UFA only to query images (UFA (Q), default) and to both query and support images (UFA (Q+S)). The results of UFA (Q+S) and UFA (Q) shown in Table \ref{tab:abl_ufa} indicate that augmenting support features results in performance degradation. This is consistent with our analysis in Section \ref{subsec23}.

Secondly, we compare our feature-level augmentation strategy with an image-level augmentation strategy, %including intra-image augmentation (\textit{i.e.}, Flip \cite{lang2022bam}) and inter-image augmentation 
\textit{i.e.}, CutMix \cite{yun2019cutmix}. In Table \ref{tab:abl_ufa}, we can see that 
%Flip improves 0.12$\%$ over baseline , 68.58$\%$) while 
CutMix leads to a 1.8$\%$ decrease compared to our default UFA (Q). 

\subsubsection{Feature reconstruction in CSM} 
We conduct the experiment to investigate different ways of feature reconstruction in CSM. By default, given a feature, we compute its weighted sum over all memory vectors to re-encode it.  
%obtain the new feature vector to replace the original feature vector.
Alternatively, we can also adopt a $k$-nearest neighbor search from a given feature and all memory vectors. In Table \ref{tab:abl_fr}, 
we report the results of $k$ being 1, 5 and all (default), 
%the idea of ``Local weighted summation'' is similar to the distance-weighted $k$-nearest neighbour algorithm \cite{Du1976DWNN}. Specifically, for each original feature vector, we first calculate its distance from all memory vectors in $\mathcal{M}$, then we choose 20 nearest-neighbour memory vectors and then calculate the weights according to the distance between these 20 nearest-neighbour memory vectors and each original feature vector, and finally calculate the weighted sum to generate the new feature vector.  
we see that the global weighted summation performs clearly better than local weighted summation.

\subsubsection{Feature reconstruction loss} 
The proposed feature reconstruction loss is important for training memory vectors (see Section \ref{subsec25}), we evaluate our model without using it (w/o $\mathcal{L}_{recon}$) in Table \ref{tab:abl_loss}. Results show that $\mathcal{L}_{recon}$ brings our model with  1.43$\%$ mIoU gain.

\subsubsection{Hyperparameter variation on CSM} 
We observe the effect of the number of memory vectors in CSM on segmentation performance. Table \ref{tab:abl_memory} shows that the best performance is achieved by 50 memory vectors. Redundant vectors are inefficient in reality.

\subsubsection{Hyperparameter variation on UFA} 
The UFA module can be flexibly inserted in any layer of the network architecture. We evaluate the inserted positions to select the optimal one for our model. Table \ref{tab:position} presents the results of six choices, among which ``B0,1,2,3'' indicates that UFA is applied after Block-$0$, Block-$1$, Block-$2$, or Block-$3$ of ResNet50.``B1+2'' achieves the highest performance.

\section{Conclusion}\label{sec6}

In this work, we propose a memory-guided network with uncertainty-based feature augmentation (MENUA) for few-shot semantic segmentation, which integrates two new modules (\textit{i.e.}, UFA and CSM) into the state of the art pipelines \cite{lang2022bam,peng2023hierarchical}. The UFA module is proposed to handle the intra-class variance by producing diversified features. The CSM module is introduced to alleviate the distribution shift between base and novel classes. It re-encodes the query features of novel classes by using the elemental information learned from base classes and stored in memory vectors. Experimental results on PASCAL-5$^i$ and COCO-20$^i$ demonstrate that our method successfully improves previous FSS models.


% \bibliographystyle{plain}
% \bibliography{main}

\end{document}